\documentclass[10pt,journal,compsoc]{IEEEtran}
\ifCLASSOPTIONcompsoc
  \usepackage[nocompress]{cite}
\else
  \usepackage{cite}
\fi
\usepackage{graphicx}
\usepackage{amsmath}
\usepackage[accsupp]{axessibility} 
\usepackage{array}
\usepackage{multirow}
\usepackage{pifont}
\usepackage{ragged2e}

\ifCLASSINFOpdf
\else
\fi
\begin{document}
%
\title{Automatic Facial Paralysis Estimation with Facial Action Units}
%
%
%
%

\author{Xuri Ge, Joemon M. Jose,
        Pengcheng Wang, Arunachalam Iyer, Xiao Liu,
        Hu Han,~\IEEEmembership{Member,~IEEE,}
\thanks{We thank Professor Brian O'reilly for sharing part of the facial paralysis dataset. 
This work was supported in part by China Scholarship Council (CSC) from the Ministry of Education of China (No. 202006310028). }
\IEEEcompsocitemizethanks{
\IEEEcompsocthanksitem Xuri Ge is with the School of Computing Science, University of Glasgow, Scotland, UK (e-mail: x.ge.2@research.gla.ac.uk).
\IEEEcompsocthanksitem Joemon M. Jose is with the School of Computing Science, University of Glasgow, Scotland, UK (e-mail: joemon.jose@glasgow.ac.uk).
\IEEEcompsocthanksitem Pengcheng Wang is with Tomorrow Advancing Life Education Group (TAL), Beijing 100080, China (e-mail: wangpengcheng2@tal.com).
\IEEEcompsocthanksitem Arunachalam Iyer is with the Department of Otolaryngology and Head and Neck Surgery, University Hospital Monklands, Airdrie, Scotland, UK, and also with University of Glasgow, Scotland, UK (aruniyerent@gmail.com).
\IEEEcompsocthanksitem Xiao Liu is with the Online Media Business Unit at Tencent, Beijing 100080, China (e-mail: ender.liux@gmail.com).
\IEEEcompsocthanksitem Hu Han is with the Key Laboratory of Intelligent Information Processing, Institute of Computing Technology, Chinese Academy of Sciences, Beijing 100190, China, and also with Peng Cheng Laboratory, Shenzhen 518055, China (e-mail: hanhu@ict.ac.cn).
}
\thanks{Manuscript received April 19, 2005; revised August 26, 2015.}}

%
%

\markboth{Journal of \LaTeX\ Class Files,~Vol.~14, No.~8, August~2015}%
{Shell \MakeLowercase{\textit{et al.}}: Bare Demo of IEEEtran.cls for Computer Society Journals}
%



\IEEEtitleabstractindextext{%
\begin{abstract}
\justifying Facial palsy is unilateral facial nerve weakness or paralysis of rapid onset with unknown cause. 
Automatically estimating facial palsy severeness can be helpful for the diagnosis and treatment of people suffering from it across the world. 
In this work, we develop and experiment with a novel model for estimating facial palsy severity. 
For this,  an effective Facial Action Units (AU) detection technique is incorporated into our model, where AUs refer to a unique set of facial muscle movements used to describe almost every anatomically possible facial expression. 
In this paper, we propose a novel \textbf{A}daptive \textbf{L}ocal-\textbf{G}lobal \textbf{R}elational \textbf{N}etwork (ALGRNet) for facial AU detection and use it to classify facial paralysis severity. 
ALGRNet mainly consists of three main novel structures: (i) an adaptive region learning module that learns the adaptive muscle regions based on the detected landmarks; (ii) a skip-BiLSTM that models the latent relationships among local AUs; and (iii) a feature fusion\&refining module that investigates the complementary between the local and global face. 
Quantitative results on two AU benchmarks, \textit{i.e.}, BP4D and DISFA, demonstrate our ALGRNet can achieve promising AU detection accuracy. 
We further demonstrate the effectiveness of its application to facial paralysis estimation by migrating ALGRNet to a facial paralysis dataset collected and annotated by medical professionals.
\end{abstract}

}

\maketitle

\IEEEdisplaynontitleabstractindextext

%
\IEEEpeerreviewmaketitle

\IEEEraisesectionheading{\section{Introduction}\label{sec:introduction}}
    \IEEEPARstart{R}{ecently}, facial action units detection has attracted increasing research attention in computer vision due to its wide range of potential applications in facial state analysis, \textit{i.e.}, diagnosing mental disease \cite{rubinow1992impaired}, face recognition \cite{5597000}, improving e-learning experiences \cite{niu2018automatic}, deception detection\cite{feldman1979detection}, \textit{etc.} On a similar line to these applications, facial palsy is affecting a large number of population and automatic estimation of facial palsy severeness can be useful for both diagnosis and treatment. 
    Facial palsy is the temporary or permanent weakness or lack of movement affecting one side of the face and is an acute, unilateral facial nerve weakness or paralysis of rapid onset (less than 72 hours) and unknown cause.
    It affects around 23 per 100,000 people per year, and current methods for facial palsy severity estimation is a relatively subjective process.
    However, in the literature it is rare to see such studies that extending the AU detection model to the assessment of facial paralysis grades. 
    In fact, the severity of facial paralysis can be estimated by the manifestations of  muscle areas of the face together, which is really similar to the representation of  individual expressions using the Facial Action Coding System (FACS) \cite{ekman1997face}. 
    In this study, we propose a novel AU detection model and explore its ability to estimate facial paralysis severity. 
    
    From a biological perspective, the activation of AU corresponds to the movement of facial muscles; however,  AU detection is challenging because of the subtle facial changes caused by AU. Hand-crafted features are used to represent the appearance of different local facial regions in early works \cite{tong2008learning,li2013simultaneous}.    
    In fact, this also applies to some works on facial state analyses, such as facial paralysis estimation \cite{cheng2010evaluation,barrios2021farapy} and patient pain detection \cite{chen2019learning}. 
    However, due to their shallow nature, hand-crafted features are not discriminative enough to depict the morphological variations in the face. 
    In order to improve the feature representation of AUs, deep learning-based approaches for AU detection have been developed in recent years.

    

\begin{figure}[t] 
	\centering
	\includegraphics[width=1\linewidth]{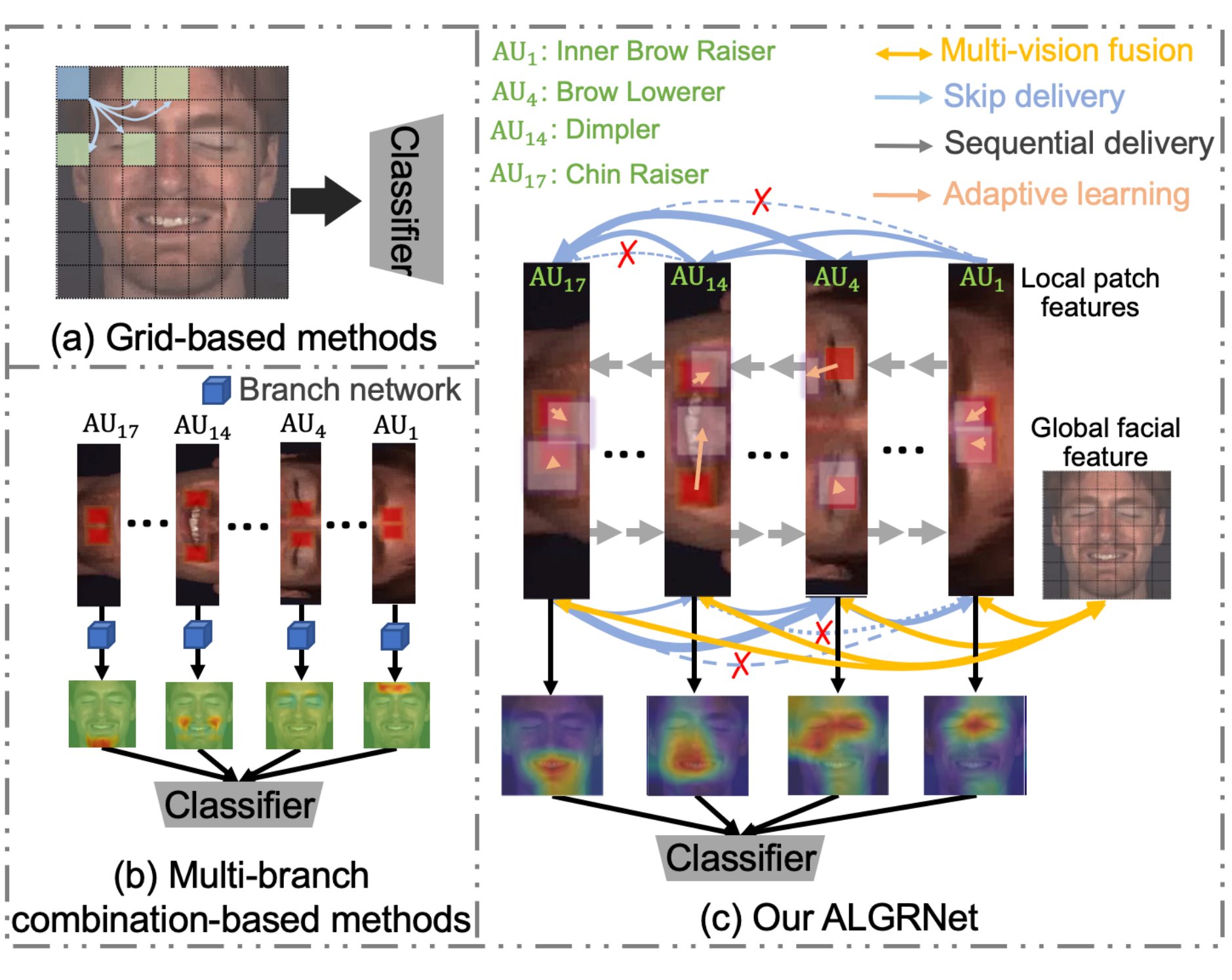}
	\caption{
	Illustration of the different schemes: (a) the traditional grid-based feature extraction and classification, (b) the popular multi-branch combination-based detection methods, and (c) our ALGRNet method. 
	Our ALGRNet, in comparison with (a) and (b), adaptively adjusts the AU areas in terms of different individuals based on detected landmarks, exploits mutual facilitation and inhibition of region-based multiple branches through a novel bidirectional structure with skipping gates and refines their irregular representations guided by the global facial feature.
    }
	\label{fig:frams}
\end{figure}

    To improve the AU representation, most existing facial AU detection methods combine local features from numerous independent AU branches, each corresponding to a separate AU patch.
    As shown in Fig. \ref{fig:frams} (a), some grid-based deep learning studies \cite{liu2014facial,liu2014feature} combine regional (patch-based) Convolutional Neural Network (CNN) features from a face with equal grids. 
    For example, LP-Net \cite{niu2019local} using an LSTM model \cite{hochreiter1997long} to combine the local CNN features from equal partition grids. 
    However, there are two significant issues with dividing the image into fixed grids; (i) It is hard to focus accurately on the muscle region related to each AU patch; and (ii) grid-based features may not adequately reflect the ROIs of irregularly shaped AU patches. 
    Recent popular multi-branch combination-based methods \cite{zhao2016joint,shao2019facial,shao2021jaa} refine the AU-related features with irregular regions by fusing global or local features from independent AU branches based on the detected corresponding muscle region, as shown in Fig. \ref{fig:frams} (b).
    For example, the scheme in \cite{shao2018deep} joints face alignment and AU detection in an end-to-end architecture, designed to extract corresponding AU features using multiple branches based on the detected and calculated AU centre coordinates. 
    Furthermore, the latest approach \cite{shao2021jaa} proposes a local AU recognition loss that refines the local attention map by the near-region pixel contributions for local regions in independent branches, rather than a pre-defined attention map; however, which also ignores the interrelationship between multiple AU areas of each face.
    
    While AU detection methods based on multi-branch combinations have shown their success in the fusion of local AU features, limitations remain in establishing interrelationships between adaptive AU regions and in modelling local and global context. 
    Firstly, unlike the localisation of corresponding AU regions based on fixed landmarks, adaptive learning of AU regions (shape and position) can improve the robustness of the model based on the diversity of expressions and individual characteristics, which  however is usually neglected by the exiting literature. 
    Secondly, according to the statistics of FACS \cite{ekman1997face}, some patches corresponding to AUs are strongly correlated in some specific expressions (here we define positive correlation (mutual assistance) if multiple patches jointly influence the activation of the target AU, otherwise negative correlation (mutual exclusion)). 
    For instance, the cheek area and the mouth corner of the face usually active simultaneously in a common facial behaviour called Duchenne smile, resulting in high correlations between AU6 (cheek raiser) and AU12 (lip corner puller).
    Furthermore, due to muscle linkage, adjacent AU2 (Outer Brow Raiser) and AU7 (Lid Tightener) will usually be activated simultaneously as the startle. 
    Inspired by these biological phenomena, we believe that it is important to capture the interactive relationship between patch-based branches, such as sequential/skipping information transfer of adjacent/non-adjacent related muscle regions, to enhance the AU features.
    Furthermore, the muscle activation areas of AU are often irregular due to individual and expression differences,  and some non-AU areas are also commonly activated due to muscle linkage.
    We therefore argue that it is vital to use the information of global faces to complement the personalisation of each AU in terms of different individuals and expressions.

    To this end, we propose a novel ALGRNet for AU detection and apply it to the facial paralysis estimation task to validate its robustness and transferability. 
    Specifically,  grid-based global features are extracted from a stem network consisting of multiple convolutional layers and
    local AU features are extracted from the calculated regions based on the detected AU centres. 
    Different from previous methods, we calculate AU centres to suit different individuals and expressions by learning the landmarks and corresponding offsets. 
    To catch the potential positive and negative relations among the branches, a skip-BiLSTM module is designed. To model the mutual support and exclusion information, the adjacent patches are  transferred in BiLSTM \cite{graves2005framewise} while the distant patches are connected via skipping-type gates. We model each branch as independent and equal and hence our kip connection manner can minimize the loss of information compared with traditional BiLSTM. Subsequently, in order to  fuse  global features to each local AU feature and also with even non-AU regional features, and in contrast to  the previous approaches \cite{shao2021jaa}, a novel gated fusion architecture in the new feature fusion\&refining module is proposed. This is important considering that the different AUs extracted from the face  may focus on different information across the face regions.
    Finally, AU features combining other beneficial AU regions as well as non-AU regions are fed into a multi-branch classification network for AU detection or facial palsy class estimation.
    
    Our contributions can be summarized as follows:
    \begin{itemize}
    \item We propose a skip-BiLSTM module to improve the robustness of local AU representations by modeling the mutual assistance and exclusion relationships of individual AUs based on the learned adaptive landmarks;
    \item We propose a feature fusion\&refining module, filtering information that contributes to the target AU, even non-AU areas, to facilitate more discriminative local AU feature generation;
    \item The proposed ALGRNet has established new state-of-the-art performance for AU detection on two benchmark datasets, \textit{i.e.,} BP4D and DISFA, without any external data or pre-trained models in additional data. 
    Notably, we exploit a facial paralysis dataset, named FPara, to verify that the proposed AU detection model can be applied to facial paralysis estimation and achieves superior performance than the baseline methods.
    \end{itemize}

    
    In comparison to the earlier conference version~\cite{ge2021local} of this work, we propose a new adaptive region learning module in Section \ref{Adaptive} in order to further improve the accuracy of muscle regions and to better adapt to irregular muscle shapes. 
    In particular, the adaptive region learning module contains learning of scaling factors to change the size of the corresponding muscle regions, as well as offset learning to slightly adjust landmark differences, with respect to different individuals. 
    This suggests that adaptive region learning could better help the model to focus more accurately on the muscle region changes corresponding to each AU, and to obtain stronger robustness and generalisation ability. 
    In addition, we did not evaluate the generalizability and transferability of the AU detection presented in the previous version, whereas we attempt it in the present study. 
    Facial paralysis estimation is a time-consuming and subjective task for a traditional physician's diagnosis. 
    Facial palsy is a condition characterised by motor dysfunction of the muscles of facial expression.
    It is usually qualified by observing the activation status of certain muscle areas and facial symmetry when the patient makes certain expressions, such as basic eyebrow raising, eye closing and mouth puckering, \textit{etc}. 
    In this study, we apply the proposed ALGRNet on facial paralysis estimation, which can improve the effectiveness of facial paralysis recognition and estimation by focusing on the activation of multiple muscle regions as well as global facial information. 
    Specifically, we exploit a facial paralysis dataset which is annotated by medical professions to four grades of facial palsy degrees, \textit{i.e.} normal, low, medium and high grade. 
    For  facial paralysis estimation we focus on the muscle areas that are more preferred in the facial paralysis ratings rather than the AU predefined muscle regions. 
    Finally, we combine the multiple muscle region features enhanced by the interaction, as well as the useful global information, to obtain the final facial features for the facial palsy grade classification. 
    To the best of our knowledge, there is no existing work in the literature on the estimation of facial paralysis using AU recognition methods. 
    And we show the effectiveness and transferability of the proposed ALGRNet quantitatively through the Facial Palsy Assessment application, which was not present in our earlier conference version \cite{ge2021local}.

\begin{figure*}[t] 
	\centering
	\includegraphics[width=1\linewidth]{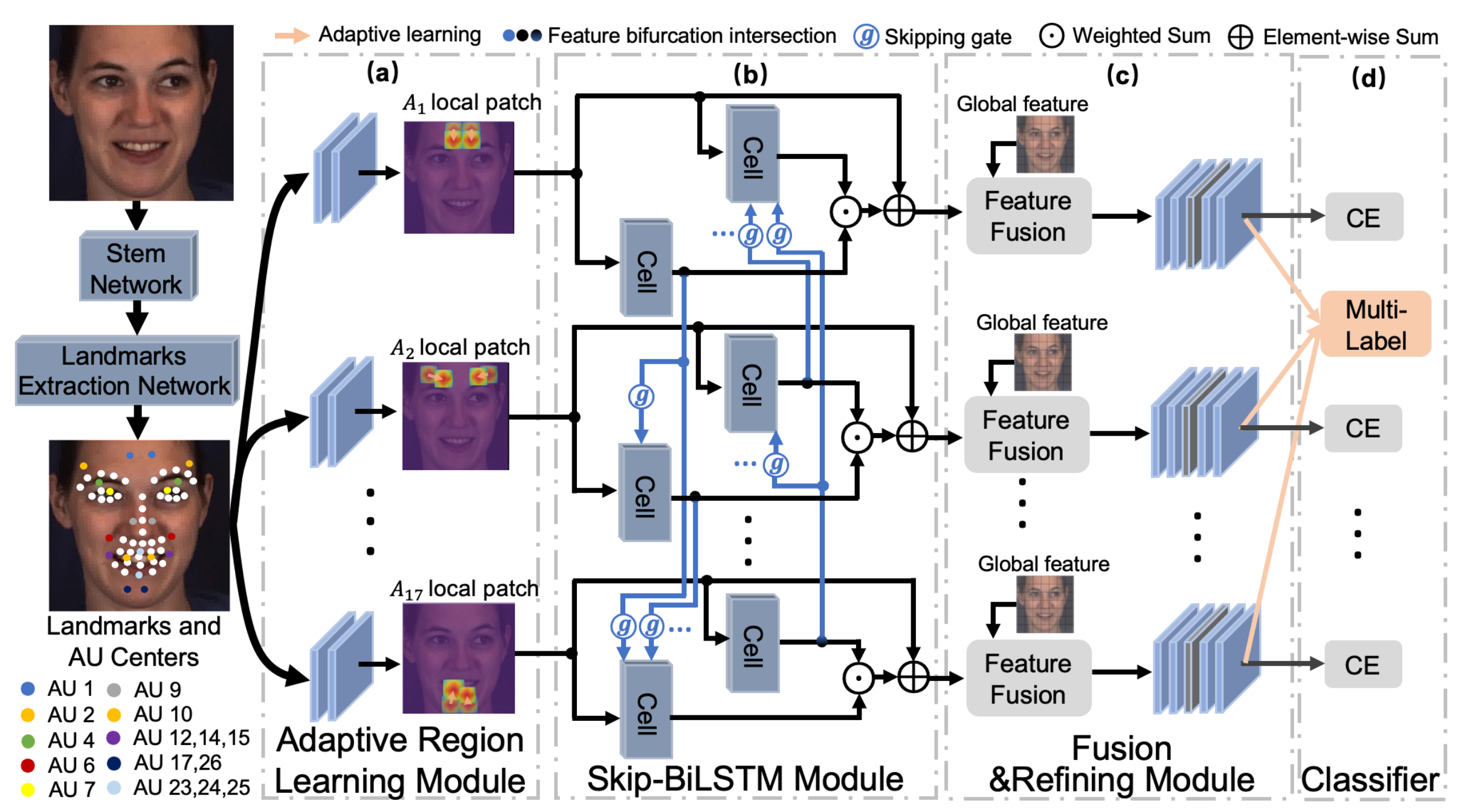}
	\caption{The overall architecture of the proposed ALGRNet for facial AU detection. 
	We utilize a simple landmark localization network to detect the landmarks and two linear-based network to learn the offsets and scaling factor of AU centers, which are used to compute local AU patches. 
	We then feed the features into the novel multi-branch network with a skip-BiLSTM module and a feature fusion\&refining module, with each branch corresponding to an AU. 
	The skip-BiLSTM module mines positive and negative relations among different AU branches by different information delivery options. 
    And the feature fusion\&refining module in each branch helps the local AU region to fit irregular shape guided by the global grid-based feature. 
	Finally, a multi-classifier is employed to predict individual AU activation probabilities. 
	}
	\label{fig:fig_overall}
\end{figure*}

\section{Related Work}
\subsection{Facial Action Units Detection}
    Automatic AU detection is a task that detects the movement of a set of facial muscles. 
    Recently, patch-learning based methods are the most popular paradigms for AU detection \cite{zhu2011dynamic, wang2013capturing, li2013data,zhao2015joint,li2017action,tang2017view, ma2019r,ntinou2021transfer}. 
    For instance, 
    \cite{taheri2014structure} used the composition rules of different fixed patches for different AUs to recover facial expressions by sparse coding. 
    \cite{jaiswal2016deep} used a CNNs and BiLSTM to extract and model the image regions for AUs, which are pre-select by domain knowledge and facial geometry. 
    %
    However, all above methods need to pre-defined the patch location first. 
    To address these issues, \cite{shao2018deep} proposed to jointly estimate the location of landmarks and the presence of action units in an end-to-end framework, where landmarks can be used to compute the attention map for each AU separately.
    Recent works \cite{tong2007facial, wu2017deep, wang2017capturing, song2018eeg, wang2018exploring, song2021hybrid} explicitly take into consideration the linkage relationship between different AUs for AU detection, which rely on action unit relationship modeling to help improve recognition accuracy. 
    Typically, \cite{tong2007facial, zhao2016deep} exploited the relationships between AU labels via a dynamic Bayesian network. 
    \cite{niu2019multi} embedded the relations among AUs through a predefined graph convolutional network (GCN). 
    \cite{li2019semantic} integrated the prior knowledge from FACS into an offline graph, which can construct a knowledge graph coding the AU correlations. 
    However, these methods require prior connections by counting the co-occurrence probabilities in different datasets in advance. 
    \cite{song2021dynamic, song2021hybrid,song2021uncertain} applied a adaptive graph to model the relationships between AUs based on global feature, ignoring local-global interactions. 

    The most relevant previous studies to ours are \cite{shao2018deep,shao2021jaa}, which combine AU detection and face alignment into a multi-branch network. 
    Different from these methods, our ALGRNet can adaptively adjust the target muscle region corresponding to each AU and utilizes the learned mutual  assistance and exclusion relationships between the target muscle and other muscle regions to enhance the feature representation of the target AU.
    Doing so allows us to provide more robustness and interpretability than \cite{shao2021jaa}. 
    
\subsection{Facial Paralysis Estimation}
 Facial paralysis estimation has recently attracted extensive research attention \cite{fields1990facial,chu2011threshold}, due to the significant psychological and functional impairment to the patients. 
 Nottingham system \cite{murty1994nottingham} is a widely accepted system for the clinical assessment of facial nerve function, which is similar with House-Brackmann (H-B) \cite{house1985facial}. 
 In addition to these, there are over twenty other methods of recognising and assessing facial palsy that are available in the literature.
 However, all of the above traditional methods are estimated by medical professionals and are both time consuming and subjective. 
 More Recently, deep learning has been widely applied for facial representation learning, including using the deep representation for face recognition, face alignment, \textit{etc}. 
 \cite{dong2008approach} and \cite{barbosa2016efficient} proposed two efficient quantitative assessment of facial paralysis based on the detected key points.
 \cite{cheng2010evaluation} proposed to obtain the facial paralysis degree by calculating the changes of the surface areas of specific facial region. 
 \cite{wang2016automatic} considered both static facial asymmetry and dynamic transformation factors to evaluate the degree of facial paralysis. 
 However, most of the exiting methods were only used the deep learning methods to pave the way for physical computation and do not directly model and predict the depth features of a face image.  In addition,
 they  apply some post-processing to obtain the final result. 
 
 In contrast to these existing methods, we employ an end-to-end deep framework ALGRNet to predict the grade of facial paralysis. 
 We perform an automated estimation by combining depth features from high interest muscle regions with feature information from the global face, without any post-processing.


\section{Approach}
    The framework of the proposed ALGRNet for AU detection is presented in Fig. \ref{fig:fig_overall}. 
    It is composed of four main modules, \textit{i.e.,} adaptive region learning module (Subsection \ref{Adaptive}) for adaptive muscle region localisation, a skip-BiLSTM module (Subsection \ref{skip-BiLSTM}) for mutual facilitation and inhibition modeling, a feature fusion\&refining module (Subsection \ref{frnet}) for refining features of irregular muscle regions, and a multi-classifier module (Subsection \ref{Overview of ALGRNet}) for predicting the AU activation probability. 
    Note that, our ALGRNet, when applied to facial palsy detection, only modifies the definition of muscle regions according to the recommendations of the physician. 
    
    \subsection{Overview of ALGRNet} \label{Overview of ALGRNet}
        Our method employs a multi-branch network \cite{corneanu2018deep,shao2018deep, shao2021jaa} for both facial paralysis estimation and facial AU detection tasks. 
        However, in contrast to previous methods, we believe that exploiting the relationship between multiple patches plays a crucial role in building a robust model for AU detection. 
        In addition, due to the diversity of expressions and individual characteristics, we also attempted to learn adaptive muscle region offsets and scaling factors for each AU region. 
        To this end, we design three modules (adaptive region learning module, skip-BiLSTM module and feature Fusion\&Refining module) based on established multi-branch network that can fully exploit inter-regional and local-global interactions on the basis of adaptively adjusted muscle localization.  
        
        We first adapt a hierarchical and multi-scale region learning network from \cite{shao2018deep} as our stem network, which is used to extract the grid-based global feature and the local muscle region features. 
        However, unlike the predefined  muscle regions based on the detected landmark in \cite{shao2018deep}, we add two simple networks combined with the previous face alignment network, named adaptive region learning module (detailed in Section \ref{Adaptive}), to adaptively learn the offsets and scaling factors for each region.
        After that, local patches $A=\{A_1,A_2,...,A_n\}$ are computed from the learned locations and their features $V=\{v_1,v_2,...,v_n\}$ can be extracted through the stem network, where $n$ is the numbers of selected patches. 
        For the sake of simplicity, we do not repeat here the detailed structure of the stem network.
        The detailed structure of the stem network is not repeated here for the sake of simplicity. 
    
        In our ALGRNet, we design a novel skip-BiLSTM module (detailed in Section \ref{skip-BiLSTM}) to address the lack of sufficient delivery of local patch information between individual branches, which can transmit information in two ways (sequential delivery and skipping delivery) on both two directions (forward and backward), in contrast to the traditional sequence spreading of LSTM.
        The sequential delivery of information enables full exploration of the contextual relationships between adjacent patches. 
        The skipping delivery highlight the interaction of information from non-adjacent related patches. 
        After skip-BiLSTM, we get a set of local patch features $S=\{s_1, s_2,...,s_n\}$, which are expected to have all the useful information from adjacent and non-adjacent AU patches. 
        
        Furthermore, a novel feature Fusion\&Refining module (detailed in Section \ref{frnet}) is developed to deal with irregular muscle areas, which can refine the local patches to obtain salient micro-level features for the global facial feature $G$. 
        Finally, the new patch-based representations $R = \{r_1, r_2, ..., r_n\}$ for AUs are obtained by integrating local muscle features and global facial features. 
        
        In this work, we integrate face alignment and face AU detection (or facial paralysis estimation) into an end-to-end learning model.
        Our goal is to jointly learn all the parameters by minimizing both face alignment loss and facial AU detection loss (or facial paralysis estimation loss) over the training set. 
        The face alignment loss is defined as:
        \begin{equation}
            \label{eq:Lalign} \mathcal{L}_{align} = \frac{1}{2d_o^2} \sum_{i=1}^{m} [(x_i-\hat{x_i})^2+(y_i-\hat{y_i})^2],
        \end{equation}
        where $(x_i,y_i)$ and $(\hat{x}_i,\hat{y}_i)$ denote the ground-truth (GT) coordinate and corresponding predicted coordinate of the $i$-th facial landmark, and $d_o$ is the ground-truth inter-ocular distance for normalization \cite{shao2021jaa}. 
        
        In this paper, we also regard facial AU detection as a multi-label binary classification task following \cite{shao2021jaa}. 
        It can be formulated as a supervised classification training objective as follows,
        \begin{equation}
            \label{eq:Lrec} \mathcal{L}_{rec} = - \frac{1}{n} \sum_{i=1}^{n} w_i [p_i {\rm log} \hat{p_i} + (1-p_i) {\rm log} (1-\hat{p}_i)],
        \end{equation}
        where $p_i$ denotes the GT probability of occurrence for the $i$-th AU, which is 1 if occurrence and 0 otherwise, and $\hat{p}_i$ denotes the predicted probability of occurrence. 
        $w_i$ is the data balance weights, which are same as in \cite{shao2018deep}. 
        Moreover, we also employ a weighted multi-label Dice coefficient loss \cite{milletari2016v} to overcome the sample imbalance problem, which is formulated as:
        \begin{equation}
            \label{eq:Ldice} \mathcal{L}_{dice} = \frac{1}{n} \sum_{i=1}^{n} w_i (1-\frac{2p_i\hat{p}_i+\tau}{p_i^2\hat{p}_i^2+\tau}),
        \end{equation}
        where $\tau$ is the smoothness parameter . Finally, the facial AU detection loss is defined as:
        \begin{equation}
            \label{eq:au} \mathcal{L}_{au} = \mathcal{L}_{rec}+\mathcal{L}_{dice},
        \end{equation} 
        Furthermore, we also minimize the loss of AU category classification $\mathcal{L}_{int}$ by integrating all AUs information, including the refined AU features and the face alignment features, which is similar to the processing of $\mathcal{L}_{au}$.
        Finally, the joint loss of our ALGRNet for facial AU detection is defined as:
        \begin{equation}
            \label{eq:Lall} \mathcal{L} = (\mathcal{L}_{au} +\mathcal{L}_{int})+ \lambda \mathcal{L}_{align}.
        \end{equation}         
        where $\lambda$ is a balancing parameter. 
        
        Similar with AU detection, we also joint the loss of facial paralysis estimation and face alignment, where the loss of facial paralysis estimation is formulated as:
         \begin{equation}
            \label{eq:palsy} \mathcal{L}_{par} = - w_i {q} {\rm Log}(\hat{q}),
        \end{equation}
        where $q$ and $\hat{q}$ are the label and predicted probability for the facial palsy grades, respectively. $w_i$ is also the data balance weights, obtained by counting the different classes in the training set. 
        Finally,  we optimize the whole end-to-end network by minimizing the jointly loss function $\mathcal{L}_{par}+ \lambda \mathcal{L}_{align}$ over the training set.

\begin{figure}[t] 
	\centering
	\includegraphics[width=0.6\linewidth]{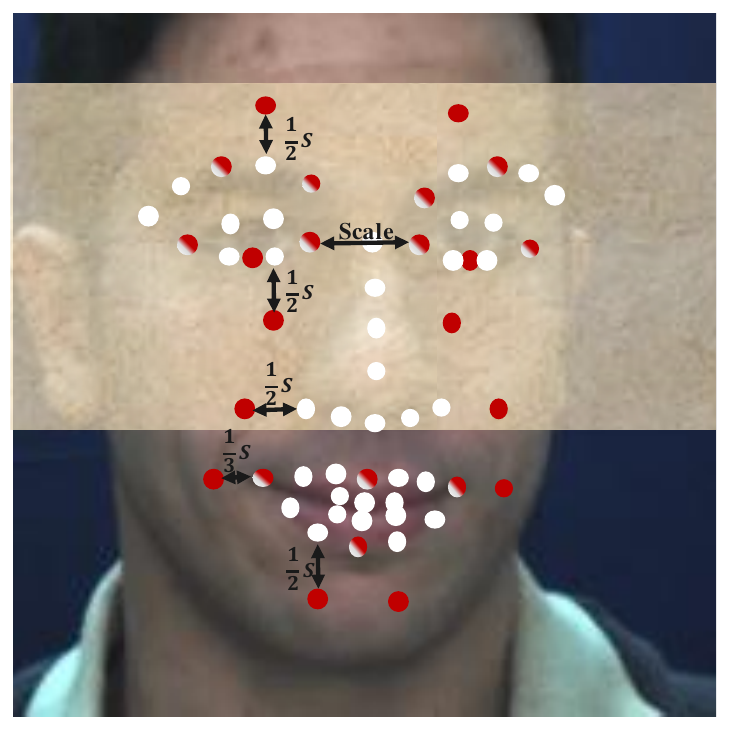}
	\caption{%
	New definitions for the 12 locations of muscle centers of facial paralysis estimation, which are marked in red or mixed red. 
	The detected landmarks are marked in white or mixed red.
	``Scale” denotes the distance between two inner eye corners.}
	\label{fig:Palsyregions}
\end{figure}        
    \subsection{Adaptive Region Learning Module} \label{Adaptive}
    Instead of the predefined muscle regions based on landmarks, we use two simple fully connected networks to adaptively learn the offsets and scaling factors for all AU regions respectively. 
    Specially, we utilize an efficient landmark extraction network after the stem network to extract the landmarks $L = \{l_1,l_2,...,l_m\}$ ($m$ is the numbers of landmarks) similar to \cite{shao2021jaa}, including three convolutional layers connected to a max-pooling layer. 
    Simultaneously, two networks containing two fully-connected layers are used to get the adaptive offsets $O = \{o_1,o_2,...,o_{2n}\}$ and scaling factors $E = \{e_1,e_2,...,e_{n}\}$ respectively. 
    According to the learned landmarks, offsets and scaling factors, local patches $A$ are calculated. 
    In particular, we first use the same rules in \cite{shao2021jaa} to get the locations of AU centers based on the detected landmarks and then update the by adding the learned offsets. 
    Please note that, we change the predefined muscle region centers, as shown in Fig. \ref{fig:Palsyregions}\footnote{Due to patient confidentiality agreements, we cannot show real patients with facial palsy. This example image is from BP4D.}, based on the detected landmarks when we apply ALGRNet on facial paralysis estimation. 
    Different from \cite{shao2021jaa}, we make the scaling factor $E$ learnable rather than a fixed value, where $e_i$ is the width ratio between the region of $AU_i$ and whole feature map. 
    After that, we generate an approximate Gaussian attention distribution for each AU region following \cite{shao2018deep}. 
    Finally, based on the learned regions, local patch features $V$ are extracted via the stem network. 
    
    \subsection{Skip-BiLSTM} \label{skip-BiLSTM}
        Fig. \ref{fig:fig_overall} (b) shows the detailed structure of our skip-BiLSTM module for contextual and skipping relationship learning.
        Specifically, we extract a set of local patch features $V=\{v_1,v_2,...,v_n\}$ from the stem network, and feed them to skip-BiLSTM. 
        Distinct from the prior works \cite{niu2019local}, we regard the multiple patches as a sequence structure from top to bottom, which can transfer information by a Bi-directional LSTM based model \cite{graves2005framewise} with our skipping-type gate. 
        Different from the traditional BiLSTM or tree-LSTM \cite{tai2015improved,ge2021structured}, our skip-BiLSTM can directly calculate the correlation between a target AU and all other AUs.
        For the $t$-th patch ($t>1$), the extracted feature $v_t$ is used to learn the weights with forward hidden states $H = \{h_1,...,h_{t-1}\}$ by the skipping-type gates, which can determine the correlation coefficient between past AUs and current AU. 
        And then the new states $\hat{H} = \{\hat{h_1},...,\hat{h}_{t-1}\}$ and $v_t$ are fed into the $t$-th forward cell in the skip-BiLSTM to learn the association weights, which can promote the transfer of relevant AUs information.
        The above process can be formulated as:
        \begin{equation}
            \overrightarrow{h_t} = {\rm Cell}(\sum_{j=1}^{t-1}{\overrightarrow{\hat{h}_j}}, v_t),
        \end{equation}
        \begin{equation}
            \overrightarrow{\hat{h}_j} = \overrightarrow{h_j}f_j,
        \end{equation}
        \begin{equation}
            f_j= {\rm \sigma}({\rm GAP}(W_j(\overrightarrow{h_j}v_t))),
        \end{equation}        
        where ${\rm Cell(\cdot)}$ indicates the basic ConvLstm cell \cite{shi2015convolutional}, and ${\rm GAP}$ denotes the global average pooling operation. 
        $W_j$ is the parameters of mapping function, in which we used ${\rm Conv2D}$.
        ${\rm \sigma}$ denotes sigmoid function. 
        We obtain the $t$-th patch feature for backward delivery, which follows the identical forward method as:
        \begin{equation}
            \overleftarrow{h_t} = {\rm Cell}(\sum_{j=t+1}^{n}{\overleftarrow{\hat{h}_j}}, v_t),
        \end{equation}
        
        In order to fully promote the information interactive among individual AUs, the final representation for each patch is computed as the average of the hidden vectors in both directions, as well as the original patch feature:
        \begin{equation}
            s_t = v_t + (\overrightarrow{h_t}+\overleftarrow{h_t})/2,
        \end{equation}

\begin{figure}[t] 
	\centering
	\includegraphics[width=0.95\linewidth]{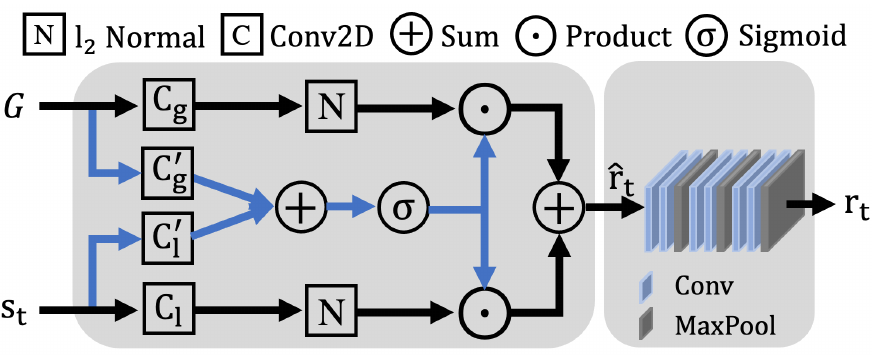}
	\caption{
	The architecture of our feature fusion\&refining module guided by global face feature.
    }
	\label{fig:FR}
\end{figure}

    \subsection{Feature Fusion\textbf{\&}Refining Module} \label{frnet}
        To exploit the useful global face feature, we design a gated fusion architecture and a refining architecture (F\&R) that can selectively balance the relative importance of local patches and global face grids.
        We add these two architectures on each local AU branch because different AUs may focus on different global information.
        The grid-based global face feature $G$ is extracted using a simple CNN with the same structure as the face alignment network \cite{shao2021jaa}. 
        As shown in Fig. \ref{fig:FR}, after obtaining the learned $t$-th local patch feature, it is fused with the grid-based global feature $G$ by the fusion architecture, which can be formulated as:
        \begin{gather}
            \alpha = {\rm {\sigma}}({\rm {C^{'}_g}} G+{\rm {C^{'}_l}} s_t),\\
            \hat{r}_t = \alpha \odot ||C_g G||_2 \oplus (1-\alpha) \odot ||C_l s_t||_2,
        \end{gather}
        where $\sigma$ is the sigmoid function, and $||\cdot||$ denotes the $l_2$-normalization. 
        ${\rm {C^{'}_*}}$ and ${\rm {C_*}}$ denote the Conv2D operation. 
        $\oplus$ denotes the element-wise weighted sum of $||C_g G||_2$ and $||C_l s_t||_2$ according to the learned gate vector $\alpha$.
        
        The final local fusion feature $s_t$ for $t$-th patch refined by our F\&R module is shown in Fig. \ref{fig:FR}. F\&R module contains three blocks. 
        Each block consists of two convolutional layers and a maxpooling layer.
        Then multi-patch features $R$ are sent to the multi-label binary classifier to calculate the occurrence probabilities of individual AUs.

\begin{table}[t]
\centering
\fontsize{9}{14}\selectfont
\setlength\tabcolsep{2pt}
\caption{Overview information of our collected facial paralysis dataset.} 
\label{tab:FAData}
\begin{tabular}{c|c|c|c|c}
\hline

\hline
Grade     & Normal      & Low       & Medium    & High \\ \hline
Num. of Video &  20         & 29        & 20        & 20     \\ \hline
Num. of Frame &  9049       & 16970     & 11019     & 10547     \\  \hline

 \hline
\end{tabular}
%
\end{table} 

\begin{table*}[t]
\centering
\fontsize{9.5}{15}\selectfont
\caption{Performance comparisons on F1-frame score of diverse AU detection for 12 AUs on BP4D. All values are in \%. * means the method employed pretrained model on additional dataset, such as ImageNet and VGGFace2, \textit{etc}, so we do not compare. The first and second places are marked with the bold font and underline, respectively.} 
\label{tab:BP4D}
\begin{tabular}{cccccccccccccc}
\hline

\hline
\multirow{2}{*}{Method} & \multicolumn{12}{c}{AU Index}                          & \multirow{2}{*}{\textbf{Avg.}} \\ \cline{2-13}
                        & 1 & 2 & 4 & 6 & 7 & 10 & 12 & 14 & 15 & 17 & 23 & 24 &                       \\ \hline
DSIN \cite{corneanu2018deep}           & \underline{51.7} & 40.4 & 56.0 & 76.1 & 73.5 & 79.9 & 85.4 & 62.7 & 37.3 & 62.8 & 38.8 & 41.6   & 58.9   \\
MLCR \cite{niu2019multi}                    & 42.4 & 36.9 & 48.1 & 77.5 & \textbf{77.6} & 83.6 & 85.8 & 61.0 & 43.7 & \textbf{63.2} & 42.1 & 55.6   & 59.8                      \\
CMS \cite{sankaran2019representation}                    & 49.1 & 44.1 & 50.3 & \textbf{79.2} & 74.7 & 80.9 & 88.2 & 63.9 & 44.4 & 60.3 & 41.4 & 51.2   & 60.6                    \\
LP-Net \cite{niu2019local}                 & 46.9 & 45.3 & 55.6 & 77.1 & 76.7 & 83.8 & 87.2 & 63.3 & 45.3 & 60.5 & \textbf{48.1} & \underline{54.2}   & 61.0                   \\
JAA-Net \cite{shao2018deep}                 & 47.2 & 44.0 & 54.9 & 77.5 & 74.6 & \underline{84.0} & 86.9 & 61.9 & 43.6 & 60.3 & 42.7 & 41.9   & 60.0              \\
ARL \cite{shao2019facial}                     & 45.8 & 39.8 & 55.1 & 75.7 & \underline{77.2} & 82.3 & 86.6 & 58.8 & \underline{47.6} & {62.1} & {47.4} & \textbf{55.4}   & 61.1                        \\

J$\rm \hat{A}$A-Net \cite{shao2021jaa}               & \textbf{53.8} & \underline{47.8} & \textbf{58.2} & \underline{78.5} & 75.8 & 82.7 & \underline{88.2} & \underline{63.7} & 43.3 & 61.8 & 45.6 & 49.9   & \underline{62.4}                   \\ \hline
UGN-B* \cite{song2021uncertain}       & 54.2 & 46.4 & 56.8 & 76.2 & 76.7 & 82.4 & 86.1 & 64.7  & 51.2 & 63.1 & 48.5 & 53.6 & 63.3  \\ 
HMP-PS* \cite{song2021hybrid}       & 53.1  & 46.1 & 56.0 & 76.5 & 76.9 & 82.1 & 86.4 & 64.8 & 51.5 & 63.0 & 49.9 & 54.5 & 63.4  \\ 
DML* \cite{wang2021dual}          & 52.6 & 44.9 & 56.2 & 79.8 & 80.4 & 85.2 & 88.3 & 65.6 & 51.7 & 59.4 & 47.3 & 49.2 & 63.4 \\\hline
                    
ALGRNet (Ours)                  & {51.2} & \textbf{48.2} & \underline{57.3} & {77.9} & {76.4} & \textbf{84.9} & \textbf{88.2} & \textbf{64.8} & \textbf{50.8} & \underline{62.8} & \underline{47.6} & 51.9   & \textbf{63.5}                     \\ \hline

\hline
\end{tabular}
\end{table*}            
\begin{table*}[t]
\centering
\centering
\fontsize{9.5}{15}\selectfont
\caption{Performance comparisons on F1-frame score of diverse AU detection for 8 AUs on DISFA. All values are in \%. The first and second places are marked with the bold font and underline, respectively. }
\label{tab:DISFA}
\begin{tabular}{cccccccccc}
\hline

\hline
\multirow{2}{*}{Method} & \multicolumn{8}{c}{AU Index}                          & \multirow{2}{*}{\textbf{Avg.}} \\ \cline{2-9}
                        & 1 & 2 & 4 & 6 & 9 & 12 & 25 & 26 &                       \\ \hline
DSIN \cite{corneanu2018deep} & 42.4 & 39.0 & 68.4 & 28.6 & 46.8 & 70.8 & 90.4 & 42.2 & 53.6 \\
CMS \cite{sankaran2019representation}               & 40.2 & 44.3 & 53.2 & \textbf{57.1} & \underline{50.3} & 73.5 & 81.1 & 59.7   & 57.4                  \\
LP-Net \cite{niu2019local}                 & 29.9 & 24.7 & \underline{72.7} & \underline{46.8} & {49.6} & 72.9 & 93.8 & 65.0   & 56.9                   \\
JAA-Net \cite{shao2018deep}                & 43.7 & 46.2 & 56.0 & 41.4 & 44.7 & 69.6 & 88.3 & 58.4   & 56.0              \\

ARL \cite{shao2019facial}                     & 43.9 & 42.1 & 63.6 & 41.8 & 40.0 & \textbf{76.2} & \textbf{95.2} & 66.8   & 58.7                  \\

J$\rm \hat{A}$A-Net \cite{shao2021jaa}         & \underline{62.4} & \underline{60.7} & 67.1 & 41.1 & 45.1 & 73.5 & 90.9 & \underline{67.4}   & \underline{63.5}            \\ \hline
 UGN-B* \cite{song2021uncertain}  & 43.3 & 48.1 & 63.4 & 49.5 & 48.2 & 72.9 & 90.8 & 59.0 & 60.0 \\
 HMP-PS* \cite{song2021hybrid} & 21.8 & 48.5 & 53.6 & 56.0 & 58.7 & 57.4 &  55.9 & 56.9 & 61.0 \\
 DML* \cite{wang2021dual} & 62.9 & 65.8 & 71.3 & 51.4 & 45.9 & 76.0 & 92.1 & 50.2 & 64.4 \\ \hline
ALGRNet                 & \textbf{63.8} & \textbf{65.4} & \textbf{73.6} & 44.5 & \textbf{54.1} & \underline{74.0} & \underline{94.7} & \textbf{69.9}   & \textbf{67.5}                     \\ \hline

\hline
\end{tabular}
%
\end{table*}  
        


\section{Experiments}

    \subsection{Dataset}
        We evaluate the effectiveness of the proposed approach for facial AU detection on popular BP4D \cite{zhang2014bp4d} and DISFA \cite{mavadati2013disfa} datasets. 
        \textbf{BP4D} consists of $328$ facial videos from 41 participants (23 females and 18 males) who were involved in 8 sessions. 
        Similar to \cite{li2017action,shao2019facial,shao2021jaa}, we consider 12 AUs and 140K valid frames with labels. 
        \textbf{DISFA} consists of 27 participants (12 females and 15 males). Each participant has a video of $4,845$ frames. 
        We also limited the number of AUs to 8 similar to \cite{li2017action,shao2021jaa}. 
        In comparison to BP4D, the experimental protocol and lighting conditions deliver DISFA to be a more challenging dataset.
        Following the experiment setting of \cite{shao2021jaa}, we evaluated the model using the 3-fold subject-exclusive cross-validation protocol.

        To evaluate the effectiveness of our ALGRNet for facial palsy severity estimation,  we exploited a facial paralysis dataset from NHS, named \textbf{FPara} (the details in Table \ref{tab:FAData}), which consists of 89 videos of facial palsy patients performing various types of facial palsy exercises inline with the House-Brackmann (H-B) scale \cite{house1985facial}.
        Each of the videos consisted of facial palsy patients performing a set of  exercises, such as raise eyebrows, close eyes gently, close eyes tightly, scrunch up face and smile, \textit{etc.} 
        They were part of a previous study on facial palsy with patient consent for research \cite{o2010objective}.  
        These videos are assigned a H-B scale from 1 to 6, and 1 being normal and 6 being severest with no body movements. 
        We then further split into four grades, such as normal (H-B score$=$1), low (H-B score$=$2), medium (3$\leq$H-B score$\leq$4) and high (5$\leq$H-B score$\leq$6) grades. FPara data is summarised in Table \ref{tab:FAData}.
        Similar to the settings of facial AU dataset, all facial paralysis grades are evaluated using subject exclusive 3-fold cross-validation, where two folds (about 80\%) are used for training and the remaining one is used for testing (about 20\%).



    \subsection{Implementation Detail}
        Our model is trained on a single NVIDIA Tesla V100 GPU with 32 GB memory.
        The whole network is trained using PyTorch \cite{paszke2019pytorch} with the stochastic gradient descent (SGD) solver, a Nesterov momentum \cite{sutskever2013importance} of 0.9 and a weight decay of 0.0005. 
        The learning rate is set to 0.01 initially with a decay rate of 0.5 every 2 epochs.
        Maximum epoch number is set to 20. 
        To enhance the diversity of training data, aligned faces are further randomly cropped into $176 \times 176$ and horizontally flipped.
        Regarding face alignment network and stem network, we set the value of the general parameters to be the same with \cite{shao2021jaa}.
        The filters for the convolutional layers in refining architecture are used $3 \times 3$ convolutional filters with a stride 1 and a padding 1. 
        In our paper, all of the mapping Conv2D operations are used $1 \times 1$ convolutional filters with a stride 1 and a padding 1. 
        The dimensionality of hidden state in ConvLstm cell is set to 64.
        The filters for the convolutional layers in ConvLstm cell are the same as refining architecture. 
        $\lambda$ is set to 0.5 for the jointly optimizing of AU detection and face alignment. 
        The ground-truth annotations of 49 landmarks of training data is detected by SDM \cite{xiong2013supervised}. 
        Different from J$\rm \hat{A}$A-Net \cite{shao2021jaa}, we averaged the predicted probability of the local information and the integrated information as the final predicted activation probability for each AU, rather than simply using the integrated information of all the AUs. 
\begin{table*}[t]
\centering
\fontsize{9}{15}\selectfont
\caption{Ablation study of ALGRNet for 8 AUs on DISFA. All values are in \%.}
\label{tab:Ablation_DISFA}
\begin{tabular}{cccc|ccccccccc}
\hline

\hline
\multirow{2}{*}{Methods} & \multicolumn{3}{c|}{Setting}                     & \multicolumn{8}{c}{AU Index}                         & \multirow{2}{*}{\textbf{Avg.}} \\ \cline{2-12}
      & S-B  & F\&R   &Ada                   & 1                 & 2                & 4                    & 6                     & 9                    & 12                & 25                 & 26              &                       \\ \hline
w/o full &         &        &            & 47.1              & 61.1             & 66.3                 & \underline{44.7}      & {52.2}     & 74.9              & 92.2               & 66.2            & 63.1                \\
w/o F\&R      &  $\surd$   &     &        & {62.6}  & 64.2             & 72.4                 & 42.3                  & 49.9                 & \textbf{76.1}     & 93.5               & \underline{72.6} & \underline{66.7}                  \\ 
w/o S-B      &        & $\surd$   &       & 58.7              & \underline{65.2}    & \underline{73.5}        & 43.9                  & \underline{53.5}        & 72.2              & {94.1}   & 64.7            & 65.7                     \\ 
w/ Bi      &    &  $\surd$     &       & 61.1             & 58.4                 & 70.9                 &{45.5}                 & 47.9     & 74.9      & 92.5  & 70.8 & 65.2                  \\ 
w/o Ada &  $\surd$ & $\surd$    &          & \underline{62.6}     & {64.4} & {72.5}     & \textbf{46.6}         & 48.8       &\underline{75.7}  & \underline{94.4}      & \textbf{73.0}    & 67.3               \\ 
\hline
ALGRNet &  $\surd$ & $\surd$    & $\surd$    & \textbf{63.8} & \textbf{65.4} & \textbf{73.6} & 44.5 & \textbf{54.1} & 74.0 & \textbf{94.7} & {69.9}   & \textbf{67.5}           \\ \hline

\hline
\end{tabular}
\end{table*}

\begin{table*}[t]
\centering
\fontsize{9}{15}\selectfont
\caption{Ablation study of ALGRNet for 12 AUs on BP4D. All values are in \%. }
\label{tab:Ablation_BP4D}
\begin{tabular}{cccc|ccccccccccccc}
\hline

\hline
\multirow{2}{*}{Methods} & \multicolumn{3}{c|}{Setting}                     & \multicolumn{12}{c}{AU Index}                         & \multirow{2}{*}{\textbf{Avg.}} \\ \cline{2-16}
      & S-B  & F\&R & Ada  & 1 & 2 & 4 & 6 & 7 & 10 & 12 & 14 & 15 & 17 & 23 & 24 &                       \\ \hline
w/o full &   &   &    & 50.1 & 47.1 & 54.3 & 77.3 & 75.1 & 82.5 & 88.1 & 61.7 & 44.9 & {62.7} & 45.2 & 49.9   & 61.6                \\
w/o F\&R      &  $\surd$   &   &       & 50.4 & 46.9 & 53.4 & \textbf{79.0} & \underline{77.4} & \underline{84.7} & 87.4 & 63.0 & 45.3 & \textbf{63.3} & {47.0} & \textbf{55.7}   & 62.8                  \\
w/o S-B      &        & $\surd$  &     & \textbf{51.3} & {47.6} & {56.3} & \underline{78.2} & 76.2 & 83.7 & 88.1 & {64.4} & 49.1 & 61.9 & 46.1 & 49.8   & 62.7                     \\
w/ Bi      &        & $\surd$  &      & 50.7 & \textbf{50.0} & 55.2 & 77.0 & 75.7 & 84.1 & \underline{88.2} & 63.4 & {49.1} & 62.3 & \underline{47.3} & 52.0   & {62.9}                     \\
w/o Ada   &  $\surd$ & $\surd$   &   & {50.8} & 47.1 & \textbf{57.8} & 77.6 & \textbf{77.4} & \underline{84.7} & \textbf{88.2} &  \textbf{66.4} & \underline{49.8} & 61.5 & 46.8 & \underline{52.3}   & \underline{63.4}         \\  \hline
ALGRNet &  $\surd$ & $\surd$   & $\surd$  & \underline{51.2} & \underline{48.2} & \underline{57.3} & {77.9} & {76.4} & \textbf{84.9} & \textbf{88.2} & \underline{64.8} & \textbf{50.8} & \underline{62.8} & \textbf{47.6} & 51.9   & \textbf{63.5}             \\ \hline

\hline
\end{tabular}
\end{table*}

    \subsection{Performance Metric.}
        We evaluate the performance of all methods in terms of the F1 score (\%) which has been widely used for classification. 
        F1-frame score is the harmonic mean of the Precision $\rm P$ and Recall $\rm R$ and calculated by $\rm F1=2PR/(P+R)$. 
        For comparison, we calculate F1 score for all facial paralysis grades on FPara and for all the AUs on DISFA and BP4D and then average them (denoted as \textbf{Avg.}) separately with ``\%” omitted. 

    \subsection{Overall Performance of AU Detection}
    We compare the proposed ALGRNet for AU detection with several single-image based baselines in Table \ref{tab:BP4D} and Table \ref{tab:DISFA}, 
    including Deep Structure Inference Network (DSIN) \cite{corneanu2018deep}, Joint AU Detection and Face Alignment (JAA-Net) \cite{shao2018deep}, Multi-Label Co-Regularization (MLCR) \cite{niu2019multi}, Cross Modality Supervision (CMS) \cite{sankaran2019representation},  Local relationship learning with Person-specific shape regularization (LP-Net) \cite{niu2019local}, Attention and Relation Learning (ARL) \cite{shao2019facial},  and Joint AU detection and face alignment via Adaptive Attention Network (J$\rm \hat{A}$ANet) \cite{shao2021jaa}. 
    The performances of the baselines in Table \ref{tab:DISFA} and \ref{tab:BP4D} are their reported results. 
    The first and second places are marked with the bold font and ``$\_$'', respectively.
    
    For a more comprehensive display, we also show methods (marked with $*$) \cite{song2021hybrid,song2021uncertain,wang2021dual} that use additional data, such as ImageNet \cite{deng2009imagenet} and VGGFace2 \cite{cao2018vggface2}, \textit{etc}, for pre-training. 
    Due to the fact that our stem network only consists of a few simple convolutional layers, even if we pre-trained on additional datasets, it is unfair compared to pre-training on deeper feature extraction networks, such as ResNet50 \cite{he2016deep}. 
    In fact, our results are still excellent compared with them, which demonstrates the superiority and effectiveness of our proposed learning scheme. 
    We omit the need for additional modal inputs and non-frame-based models \cite{yang2020adaptive,liu2019multi}. 

    \textbf{Quantitative comparison on BP4D:}  
    We report the performance comparisons between our ALGRNet and baselines on BP4D in Table \ref{tab:BP4D}. 
    As it can be observed, our ALGRNet significantly outperforms all the other methods in terms of F1-frame score and achieves the fist and second places for most of the 12 AUs annotated in BP4D. 
    J$\rm \hat{A}$ANet is the latest state-of-the-art method which also joint AU detection and face alignment into an end-to-end multi-label multi-branch network. 
    Our ALGRNet achieves 1.1\% higher average F1-frame score compared with J$\rm \hat{A}$A-Net. 
    The main reason lies in our ALGRNet overcomes the problem of non-transferable information between branches in the J$\rm \hat{A}$A-Net and adaptively adjusts the muscle regions corresponding to the AU.

    \textbf{Quantitative comparison on DISFA:} 
    We also report the performance of our proposed ALGRNet on DISFA. 
    Table \ref{tab:DISFA} shows the performance of our ALGRNet is the best in terms of average F1 score compared with all baselines.
    And our approach significantly outperforms all other methods for most of the 8 AUs annotated in DISFA. 
    Compared with the existing end-to-end feature learning and multi-label classification methods DSIN \cite{corneanu2018deep} and ARL \cite{shao2019facial}, the average F1-frame score of our proposed ALGRNet get 13.9\% and 8.8\% higher, respectively.
    Moreover, compared with the multi-branch combination-based state-of-the-art method J$\rm \hat{A}$ANet \cite{shao2021jaa}, our ALGRNet achieves 4.0\% improvements in terms of average F1-frame score. 
    
    The eventual experimental results of ALGRNet demonstrate its effectiveness in improving AU detection accuracy on DISFA and BP4D, as well as good robustness.

    \subsection{Ablative Analysis} 
    To fully examine the impact of our proposed adaptive region learning module, skip-BiLSTM module and feature fusion\&refining module, we conduct detailed ablative studies to compare different variants of ALGRNet for facial AU detection on DISFA and BP4D. 
    
    \subsubsection{Effects of adaptive region learning module}
    To cancel out the adaptive region learning (indicated w/o Ada), we fellow the same experiment setting as \cite{shao2021jaa} (It means each scaling factor $e$ is set to 0.14.) to predefined muscle region based on the detected landmarks for each AU. 
    In Table \ref{tab:Ablation_DISFA} and \ref{tab:Ablation_BP4D}, ALGRNet decreases its F1 score to 67.3\% and 63.4\% on DISFA and BP4D respectively. 
    It has been shown that the adaptive region learning module can contribute the AU detection greatly. 
    
    \subsubsection{Effects of skip-BiLSTM}
    In Table \ref{tab:Ablation_DISFA} and \ref{tab:Ablation_BP4D}, when the skip-BiLSTM module is removed (indicated by w/o S-B), ALGRNet (without adaptive region learning module) shows an absolute decrease of 1.6\% and 0.7\% in the average F1-frame score for DISFA and BP4D, respectively.
    In addition, in order to explicitly validate the effectiveness of our skipping operation, we use the basic BiLSTM \cite{graves2005framewise} (indicated by w/ Bi) instead of skip-BiLSTM for information sequential transfer across different branches in the ALGRNet (also with Fusion\&Refining module), ALGRNet obtains lower average F1 scores of 65.2\% and 62.9\% on DISFA and BP4D, respectively.
    The performance reduction clearly verifies that roughly defining the relationships between AU-related branches from top to bottom may not be the best way to model the real relationships between AUs.
    Notably, skipping operation can significantly improve performance, suggesting that our skip-type gates play an important role in our model. 
    Furthermore, the eventual experimental results demonstrate the effectiveness of our ski-BiLSTM in modelling mutual assistance and exclusion relationship between different patch-based branches the for AU detection 
    
\begin{table}[t]
\centering
\fontsize{9}{15}\selectfont
\caption{Mean error (lower is better) results of different face alignment models on BP4D, DISFA and FPara. All values are in \%.}
\label{tab:landmark}
\begin{tabular}{c|ccc}
\hline

\hline
Methods & BP4D  & DISFA & FPara \\ \hline
J$\rm \hat{A}$A-Net  & 3.80     &  3.87   &  \textbf{5.15}   \\ 
ALGRNet  & \textbf{3.78}      & \textbf{3.29}   &  5.18   \\ 
\hline

\hline
\end{tabular}
\end{table}

    \subsubsection{Effects of feature fusion\&refining module} 
    In order to illustrate the effectiveness of feature fusion\&refining module, we directly conduct the classification over the output of skip-BiLSTM without fusion\&refining module (indicated by w/o F\&R in Table \ref{tab:Ablation_DISFA} and \ref{tab:Ablation_BP4D}).
    When the fusion\&refining module is not used, the average F1-frame score drops significantly from 67.3\% to 66.7\% on DISFA and from 63.4\% to 62.8\% on BP4D, due to the lack of supplementary information from the global face for each patch.
    This suggests that the refined local AU features from the proposed fusion\&refining module, guided by the grid-based global features, make a significant contribution in our model. 
    
    Finally, after simultaneously removing all the proposed adaptive region learning module, skip-BiLSTM and fusion\&refining module (marked by w/o full in Table \ref{tab:Ablation_DISFA} and \ref{tab:Ablation_BP4D}), a significant performance degradation in AU detection can be observed, \textit{i.e.}, a 4.4\% drop on DISFA and a 1.9\% drop on BP4D in terms of average F1-frame score.
    This sufficiently demonstrates that the potential mutual assistance and exclusion relationships between the adaptive AU patches, complemented by the global facial features, can significantly improve the performance of facial AU detection. 

    \subsection{Results for Face Alignment}
    We integrate face alignment and face AU detection into our end-to-end ALGRNet, which can be beneficial for each other as they are coherently related to each other. 
    For example, the detected landmarks can help the model focus on the exact muscle areas of the AU patches. 
    As shown in Table \ref{tab:landmark}, compared with baseline method J$\rm \hat{A}$A-Net \cite{shao2021jaa}, our ALGRNet performs comparably to baseline on FPara and better on BP4D and DISFA. 
    The robustness of the adaptive region learning module allows our ALGRNet to outperform J$\rm \hat{A}$A-Net in AU detection and facial paralysis estimation, even if sometimes with slightly lower landmark detection accuracy. 
   
\begin{figure}[t] 
	\centering
	\includegraphics[width=1\linewidth]{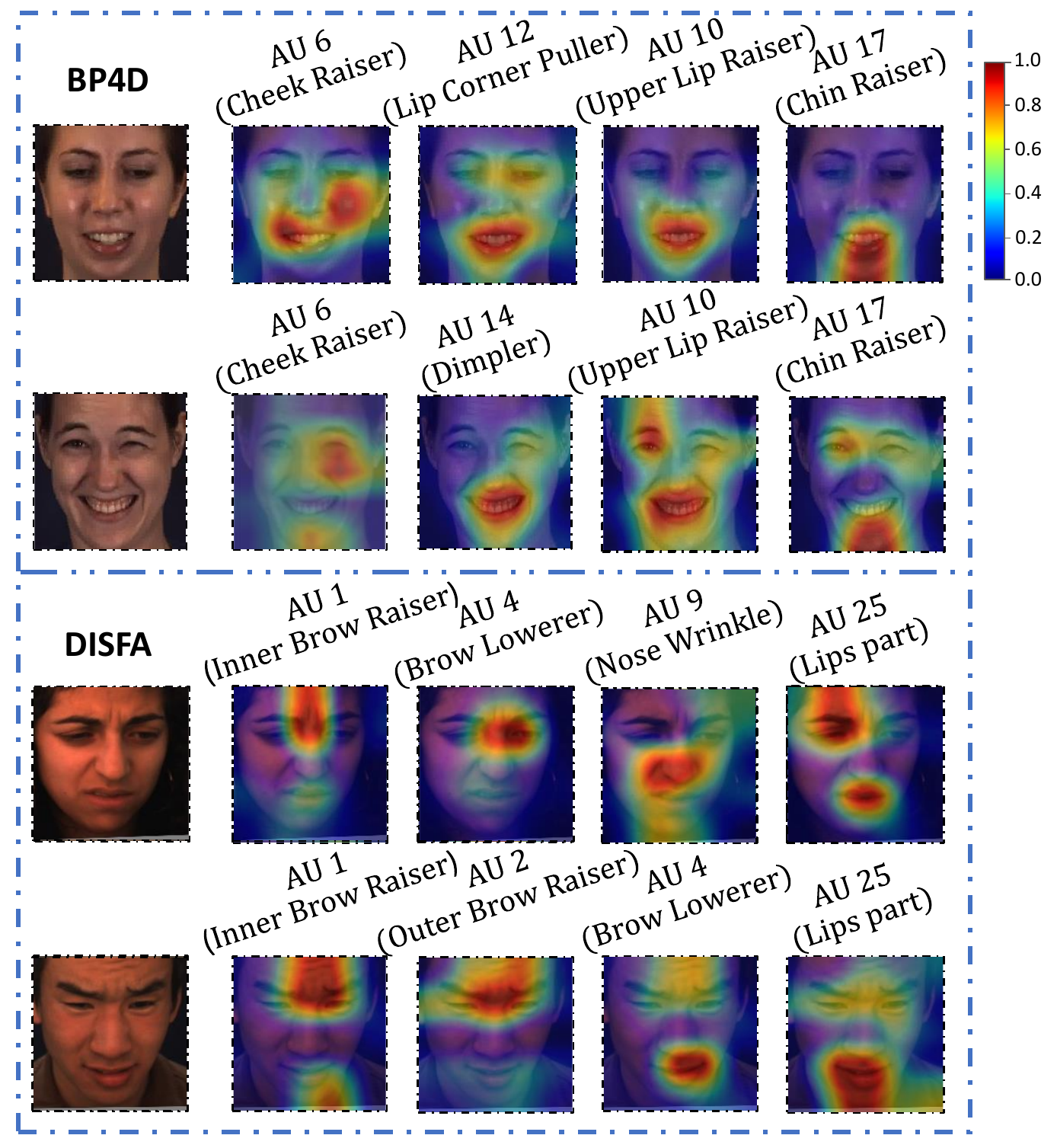} 
	\caption{Class activation maps that show the discriminative regions for different AUs in terms of different expressions and individuals on DISFA and BP4D datasets.}
	\label{fig:visual}
\end{figure}  

    \subsection{Visualization of Results}
    Fig. \ref{fig:visual} shows some examples of the learned class activation maps of ALGRNet (the outputs of F\&R module) corresponding to different AUs. 
    For an adequate display, we show two individuals from BP4D and DISFA respectively, containing visualisations of different genders with different AU categories.
    Through the learning of ALGRNet, not only the concerned AU regions can be accurately located, but also the positive (in red) or negative (in blue) correlation with other AU areas can be established and other details of the global face can be supplemented. 
    This obviously improves the flaws of the excessive localisation of J$\rm \hat{A}$A-Net \cite{shao2021jaa} and the negative influence of unrelated regions of ARL \cite{shao2018deep}. 
    In addition, it also adapts well to irregular muscle areas for different AUs.
    The heatmaps for the same AU category in the different examples are broadly consistent but also vary slightly by individual, demonstrating that our ALGRNet can learn certain rules across different datasets and adaptively adjust to different samples. 
    

\subsection{Facial Palsy Severity Estimation}
    In this section we evaluate the effectiveness of facial palsy severity estimation.
    
\begin{table}[t]
\centering
\fontsize{9}{14}\selectfont
\setlength\tabcolsep{2pt}
\caption{Performance comparisons on F1-frame score (in \%) of diverse facial paralysis estimation for 14 grades on FPara.}
\label{tab:Paralysisdata}
\begin{tabular}{cccccc}
\hline

\hline
\multirow{2}{*}{Method} & \multicolumn{4}{c}{Facial Paralysis Grades} & \multirow{2}{*}{\textbf{ Avg.}} \\ \cline{2-5}
                                 & Normal     & { Low }     & Medium     & High    &                       \\ \hline
ResNet18  \cite{he2016deep}      & 99.8       & 50.7    & 47.7       &  67.9   &   66.5                    \\
ResNet50  \cite{he2016deep}      &  99.9      & 53.9    &  54.7       & 71.4    &  70.0                   \\
Transformer-based\cite{valanarasu2021medical}  &  100   & \textbf{63.0}   & 58.6   & 68.7     & 72.6                          \\
J$\rm \hat{A}$A-Net \cite{shao2021jaa}     &  \underline{100}    & \underline{58.8}   & \underline{64.3}   &  \underline{72.9}    & \underline{72.8}                       \\ \hline
ALGRNet(Ours)                    & \textbf{100}     & 55.9  & \textbf{72.1} &  \textbf{73.2}  &  \textbf{75.4}                   \\ \hline

\hline
\end{tabular}
%
\end{table}    
\subsubsection{Facial paralysis estimation}
    Different from facial AU detection, the exiting deep-learning-based facial paralysis estimation methods are rare, so we apply currently popular deep learning classification methods, such as the ResNet \cite{he2016deep} and Transformer\cite{vaswani2017attention}, on our collected facial palsy dataset (FPara). 
    Besides, we also compare it with the state-of-the-art AU detection approach, J$\rm \hat{A}$A-Net \cite{shao2021jaa}.
    Specially, we evaluate the following methods: 
    \begin{itemize}
    \item ResNet18 and ResNet50 \cite{he2016deep}: These methods use different depth layers based on ResNet to model the input face images, which are similar to \cite{song2018neurologist}.
    \item Transformer-based method \cite{valanarasu2021medical}: This baseline is motivated from self-attention and uses the Transformer \cite{vaswani2017attention} architecture. 
    The output of the Transformer-based encoder \cite{valanarasu2021medical} is treated as the latent representation for the input of the multi-label AU classifier.
    \item J$\rm \hat{A}$A-Net \cite{shao2021jaa}: This is a recently proposed multi-branch combination-based AU detection method, which can extract precise local muscle features thanks to a joint facial alignment network.
    \end{itemize}
     
    \subsubsection{Quantitative comparison on the collected FPara} 
    Facial paralysis estimation results by different methods on our FPara are shown in Table \ref{tab:Paralysisdata}. 
    It has been shown that our ALGRNet outperforms all its competitors with impressive margins. 
    Specifically, compared with the state-of-the-art AU detection method J$\rm \hat{A}$A-Net \cite{shao2021jaa}, our ALGRNet achieves 2.6\% improvements in terms of average F1 score. 
    Moreover, the average F1 score of our ALGRNet get 2.8\% higher compared to the currently popular Transformer-based approach \cite{valanarasu2021medical}.

    The eventual experimental results of our ALGRNet demonstrate that it is successful in boosting AU detection accuracy on BP4D and DISFA, as well as having high generalisation ability on our new facial paralysis dataset.

\section{Conclusion}
    In this paper, we present ALGRNet, a novel adaptive local-global relational network for both facial action units detection and facial paralysis severity  estimation, which can exploit the mutual assistance and exclusion relationships of adaptive muscle regions as well as interactions with global information. 
    Specifically, ALGRNet adaptively represents the corresponding muscle areas in terms of different expressions and individual characteristics. 
    It then models the potential mutual assistance and exclusion relationships between AU branches and enables efficient information transfer via a novel skip-BiLSTM to get local features. 
    Finally, a novel feature fusion$\&$refining module is proposed and equipped in each branch, facilitating complementarity between local feature and grid-based global feature as well as adaptation to irregular muscle regions.
    Experimental results on two widely used AU detection benchmarks show the effectiveness of the AU detection  and the superiority of the Facial Palsy severity estimation on a  facial paralysis estimation benchmark. This not only witnesses the distinct performance gains over the state-of-the-arts, but also demonstrates effective migration capability from AU detection to the facial paralysis estimation task.

\bibliographystyle{IEEEtran}
\bibliography{egbib}

%

\begin{IEEEbiography}[{\includegraphics[width=1.2in,height=1.25in,clip,keepaspectratio]{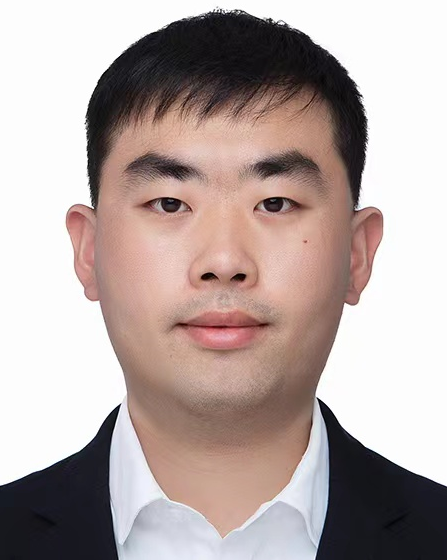}}]{Xuri Ge} received the M.S. degree in computer science from the School of Information
Science and Engineering, Xiamen University, China, in 2020.
He is currently pursuing the Ph.D. degree with the school of computer science, University of Glasgow, Scotland, UK.
His current research interests include computer vision, medical image processing, and multimedia processing.
\end{IEEEbiography}
\begin{IEEEbiography}[{\includegraphics[width=1.2in,height=1.25in,clip,keepaspectratio]{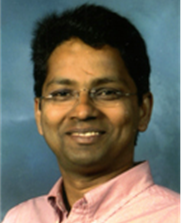}}]{Joemon M. Jose}  is a Professor at the School of Computing Science, University of Glasgow, Scotland and a member of the Information Retrieval group. His research focuses around the following three themes: (i) Social Media Analytics; (ii) Multi-modal interaction for information retrieval; (iii) Multimedia mining and search. He has published over 300 papers with more than 8000 Google Scholar citations, and an h-index of 47. He leads the Multimedia Information Retrieval group which investigates research issues related to the above themes.
\end{IEEEbiography}
\begin{IEEEbiography}[{\includegraphics[width=1.2in,height=1.25in,clip,keepaspectratio]{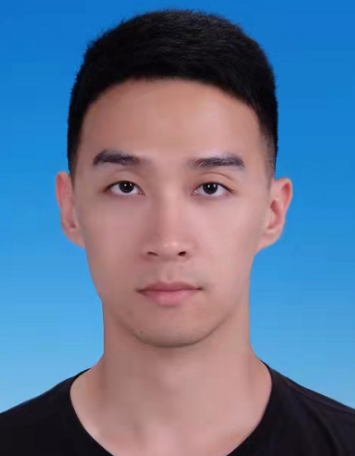}}]{Pengcheng Wang}  is currently a Researcher with Tomorrow Advancing Life Education Group (TAL), Beijing, China. His research interests include the applied AI, such as intelligent multimedia processing, and computer vision. As a Key
Team Member, he achieved the best performance in various competitions,
such as  the EmotioNet facial expression recognition challenge. 
\end{IEEEbiography}
\begin{IEEEbiography}[{\includegraphics[width=1.2in,height=1.25in,clip,keepaspectratio]{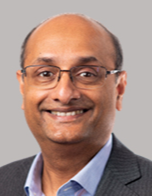}}]{Arunachalam Iyer} is a Honorary Associate Clinical Professor of Surgery, University of Glasgow, Scotland, UK. 
He is also a consultant ENT surgeon, University Hospital Monklands, UK. 
He is currently working as a Consultant ENT surgeon \& Otologist at Lanarkshire and a council member of the British society of Otology and the section of Otology, Royal society of Medicine. 
He also teaches at the Temporal bone course at University of Glasgow and is involved in training junior doctors.
\end{IEEEbiography}
\begin{IEEEbiography}[{\includegraphics[width=1.2in,height=1.25in,clip,keepaspectratio]{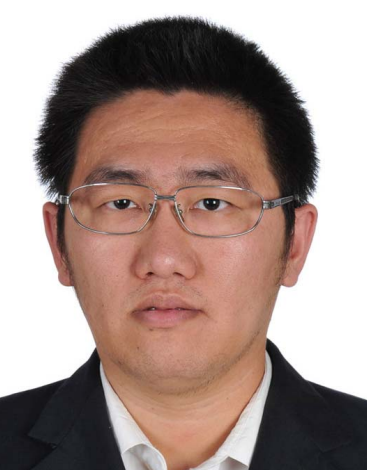}}]{Xiao Liu} received the Ph.D. degree in computer science from Zhejiang University in 2015. He worked at Baidu from 2015 to 2019 and at Tomorrow Advancing Life Education Group (TAL) from 2019 to 2021. He is currently a Researcher with the Online Media Business Unit at Tencent, Beijing, China. His research interests include the applied AI, such as intelligent multimedia processing, computer vision, and learning systems. His research results have expounded in more than 40 publications at journals and conferences, such as IEEE TRANSACTIONS ON IMAGE PROCESSING, IEEE TRANSACTIONS ON NEURAL NETWORKS AND LEARNING SYSTEMS, CVPR, ICCV, ECCV, AAAI, and MM. As a Key Team Member, he achieved the best performance in various competitions, such as the ActivityNet challenges, NTIRE super resolution challenge, and EmotioNet facial expression recognition challenge. 
\end{IEEEbiography}
\begin{IEEEbiography}[{\includegraphics[width=1.2in,height=1.25in,clip,keepaspectratio]{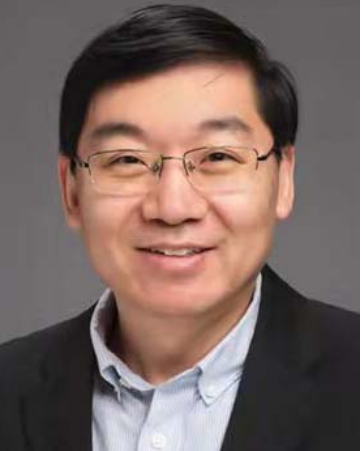}}]{Hu Han} (Member, IEEE) is a Professor at the Institute of Computing Technology (ICT), Chinese Academy of Sciences (CAS). His research interests include computer vision, pattern recognition, biometrics, and medical image analysis. He has published more than 70 papers with more than 4700 Google Scholar citations. He was a recipient of the IEEE Signal Processing Society Best Paper Award (2020), ICCV2021 Human-centric Trustworthy Computer Vision Best Paper Award, IEEE FG2019 Best Poster Presentation Award, and 2016/2018 CCBR Best Student/Poster Award. He is/was the Associate Editor of Pattern Recognition, IJCAI2021 SPC, ICPR2020 Area Chair, ISBI2022 Session Chair, and VALSE LAC. He has organized a number of special sessions and workshops of “vision based vital sign analysis and health monitoring” in ICCV2021, CVPR2020, FG2019/20/21.
\end{IEEEbiography}

\end{document}